# A Unified View-Graph Selection Framework for Structure from Motion

Rajvi Shah        Visesh Chari        P J Narayanan
Center for Visual Information Technology, IIIT Hyderabad, India
{rajvi.shah@research.iiit.ac.in, visesh@gmail.com, pjn@iiit.ac.in}

## Abstract

*View-graph is an essential input to large-scale structure from motion (SfM) pipelines. Accuracy and efficiency of large-scale SfM is crucially dependent on the input view-graph. Inconsistent or inaccurate edges can lead to inferior or wrong reconstruction. Most SfM methods remove 'undesirable' images and pairs using several, fixed heuristic criteria, and propose tailor-made solutions to achieve specific reconstruction objectives such as efficiency, accuracy, or disambiguation. In contrast to these disparate solutions, we propose a single optimization framework that can be used to achieve these different reconstruction objectives with task-specific cost modeling. We also construct a very efficient network-flow based formulation for its approximate solution. The abstraction brought on by this selection mechanism separates the challenges specific to datasets and reconstruction objectives from the standard SfM pipeline and improves its generalization. This paper demonstrates the application of the proposed view-graph framework with standard SfM pipeline for two particular use-cases, (i) accurate and ghost-free reconstructions of highly ambiguous datasets using costs based on disambiguation priors, and (ii) accurate and efficient reconstruction of large-scale Internet datasets using costs based on commonly used priors.*

## 1. Introduction

View-graph is a crucial input structure in context of large-scale structure from motion (SfM). Nodes in this graph represent images (also referred to as cameras or views) and edges represent relative motion or epipolar geometries (EGs) between nodes they connect. View-graphs help in 'organizing' unordered image collections useful to select, (i) a core set of images for reconstruction, and (ii) identify noisy EGs that might degrade the quality of reconstruction. State-of-the-art SfM methods like incremental [20, 29, 17], hierarchical [4, 8, 1], or global [19, 3, 14, 13, 2], all rely on a view-graph based initial step for ordering and pruning images for efficient and accurate reconstruction.

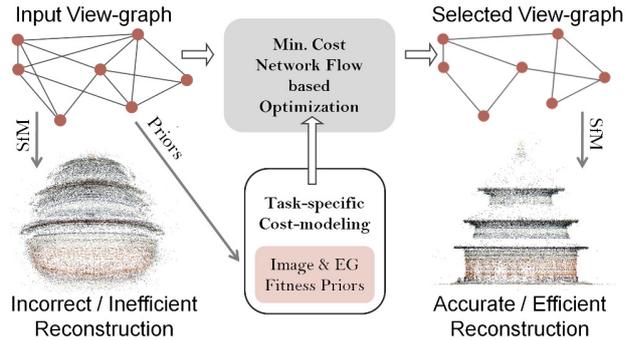

Figure 1: Outline of the proposed unified framework. With appropriately modeled costs, the framework can select view-graphs that meet desired reconstruction objectives.

Large community photo collections often display point-of-view bias, as some viewing angles are more popular than others. Using the full view-graph for SfM is often computationally expensive and unnecessary due to the high redundancy in the image set. Moreover, for closely clustered images, such narrow baseline pairs increase the uncertainty of triangulation, causing large re-projection errors. While isolated erroneous EGs can be overcome by robust averaging (global SfM) and repeated bundle adjustment (incremental SfM), large number of incorrect or ill-conditioned EGs can result in incorrect reconstruction. Hence, it is also crucial to select 'good' images and more importantly 'good' pairs for accurate reconstruction. Standard SfM methods apply fixed heuristics such as inliers, baselines, homography, loop closure for conservative selection and pruning of images and pairs. While these heuristics work well for a large variety of datasets, they are insufficient to identify and remove consistent noise that arise in highly ambiguous scenes.

Ambiguity in pairwise matching arises due to the fact that man-made structures often comprise of repetitions (windows, pillars, arches), symmetries (circular structures, identical facades), and large duplicate instances (minarets, domes). While standard SfM pipelines are robust enough to handle large number of inaccurate EGs in isolation, for scenes with high ambiguity, such wrong EGs form large consistent sets, causing ghosting errors in form of misregistered cameras, 'phantom' structures, or structures incorrectly folded along symmetry. Previous approaches propose tailor-made SfM solutions to handle such difficult scenes with local and global steps for 'reasoning' symmetry and ambiguity [32, 16, 12, 26, 11, 30]. While these methods show promising results on some very challenging datasets, they do not operate within the framework of traditional SfM pipelines. We believe that many of these challenges can be addressed under the same umbrella of 'selecting' a complete, consistent and noise-free view-graph to achieve accurate and efficient reconstruction.

The main motivation of our work is to formalize the core problem of view-graph selection and introduce a method to meet different subgraph selection objectives within a unified framework. We pose this as an optimization problem with image and pair selection costs and propose a novel network-flow based formulation for its efficient approximate solution. This abstraction allows different objectives to be achieved by plugging in task-specific costs while keeping the overall selection and reconstruction framework the same. Figure 1 shows an outline of this framework. We demonstrate the utility of our framework to successfully meet two reconstruction objectives with different challenges, (i) accurate and ghost-free reconstructions of highly ambiguous datasets, and (ii) accurate and efficient reconstruction of large-scale Internet landmarks datasets. For disambiguation task, we show how pairwise selection cost can be modeled using geometric and context priors, and introduce a new context prior. For Internet datasets reconstruction, we show how commonly used heuristics can be used to model image and pairwise selection costs.

To the best of our knowledge, this is the first attempt at systematizing view-graph selection. The proposed framework brings greater flexibility and generalization to standard SfM pipelines and its application is not limited to the specified use-cases. A unified framework also lands optimal view-graph selection as a learning problem if and when task-specific ground-truth data becomes available.

## 2. Related Work

In SfM literature, the view-graph selection problem is generally not dealt with in a unified sense. Instead, algorithms are often designed to meet specific objectives for subgraph selection. For efficiency and completeness goals, incremental/hierarchical SfM methods compute spanning subgraphs [22, 15, 9, 1], with different criteria and selection strategies. For global methods, optimality of a subgraph is often related to consistency and robustness goals [13, 31, 6, 25, 14, 28]. Our framework is not a one-size-fits-all replacement to these specialized methods, but provides a unified mechanism to express different objectives using task-specific image and pairwise selection costs. Exploring all possible objectives that can be addressed in this framework is beyond the scope of this work, in this section, we focus on the prior works that relate to the general objective of accuracy and the more specific objective of disambiguation, and discuss their relevance to our work.

### 2.1. Implicit view-graph filtering in SfM methods

Pruning undesirable images and pairs from the input view-graph is an implicit part of both incremental and global SfM methods and is often done by putting thresholds on the various criteria. Accuracy of incremental SfM hinges on seed pairwise reconstruction and next best view selection. Wide baseline seed pair selection is ensured using criteria based on epipolar inliers (that don't fit a homography) [20, 21], or inlier ratios (to detect pairwise motion as planar, rotational, or general) [17]. For next best view selection, triangulation angle, inliers, or correspondence distribution statistics are used as criteria [20, 17].

Global SfM methods first use relative rotations to estimate global rotations and then estimate global translations [19, 3, 13, 2, 27, 25]. In [3], to reduce the state space of camera parameters for MRF based estimation, relative twist ($> 20°$) and unusual aspect ratios are used as view filtering criteria. Methods that use lie-algebraic averaging of relative rotations [5, 6, 31, 14, 13, 2, 25] often discard images with unknown calibration and handle outlier EGs with loop consistency constraints [31, 14, 13], and robust cost functions.

Many of the heuristics and criteria discussed above are easy to incorporate as image and pairwise selection costs into our framework. Other heuristics that employ global reasoning (such as loop consistency) can be potentially remodeled as a pairwise heuristic.

### 2.2. SfM methods for Disambiguation

Initial methods for disambiguation focused on inferring missing correspondences [32] and using it as a prior with global objectives of camera pose inference [16] or consistent view-graph expansion [12]. Other methods consist of social network principles based track refinement [26], triplet-consistent graph matching and expansion [18], and geodesic manifold based ambiguity detection and correction [30]. A post-reconstruction approach to disambiguation uses back-projected 3D points to identify conflicting observations [10, 11]. In contrast to these specialized algorithms, we use the proposed view-graph selection framework along with a combination of disambiguation priors for cost modeling. We also propose a new context based pairwise selection prior that is based on both, missing correspondences and conflicting observations and unlike [10], this prior can be computed before reconstruction. Our solution is more general and efficient as compared to prior methods and still achieves good reconstructions on a variety of ambiguous datasets.

## 3. Notation and Problem Formulation

Let the input view-graph be $\mathcal{G} = (\mathcal{V}, \mathcal{E})$, where the set of vertices (nodes) $\mathcal{V}$ represents the images and the set of edges $\mathcal{E}$ represents the pairwise epipolar geometries (EGs). The goal is to select a subset of nodes $\mathcal{V}'$ and a subset of edges $\mathcal{E}'$ (a subgraph $\mathcal{G}' = (\mathcal{V}', \mathcal{E}')$) that meets a desired objective. We denote the indicator variables for image selection as $\delta_i$ and pair selection as $\delta_{ij}$. Here, each $\delta_i$ corresponds to the image vertex $v_i \in \mathcal{V}$ and each $\delta_{ij}$ corresponds to the pairwise edge $e_{ij} \in \mathcal{E}$. The selection problem can be represented as a minimization of the following form,

$$\begin{aligned}
\arg\min \quad & f(\delta) = \sum_i c_i \delta_i + \sum_i \sum_j c_{ij} \delta_{ij} \\
\text{subject to} \quad & \sum \delta_i \leq N, \sum \delta_{ij} \leq M \\
& i \in \mathcal{V}, (i,j) \in \mathcal{E}, \delta_i, \delta_{ij} \in \{0, 1\}.
\end{aligned} \quad (1)$$

Activation of indicator variables $\delta_i$ and $\delta_{ij}$ imply selection of corresponding image $v_i$ and pair $e_{ij}$. $c_i$ is the cost of selecting the view $v_i$ and $c_{ij}$ is the cost of selecting $e_{ij}$. Modeling these costs appropriately can express different objectives for view-graph selection. In this paper, we discuss cost modeling for accurate and ghost-free reconstructions using a few relevant priors. However, this formulation can be used to achieve other objectives using a variety of other priors known in the literature or even learned costs.

This representation is slightly problematic, since selection of images and pairs is not independent. Since, each image is most likely connected to more than one image, activation of selection variables $\delta_i$ and $\delta_j$ does not necessarily imply selection of the image pair $(i, j)$. On the contrary, selection of image pair $(i, j)$ implies that both images $i$ and $j$ constituting the pair must be selected. We model this dependency by adding additional constraints to the minimization.

$$\begin{aligned}
\arg\min \quad & f(\delta) = \sum_i c_i \delta_i + \sum_i \sum_j c_{ij} \delta_{ij} \\
\text{subject to} \quad & \sum \delta_i \leq N, \sum \delta_{ij} \leq M \\
& \forall i, j \; \delta_{ij} \leq \delta_i, \; \delta_{ij} \leq \delta_j, \\
& i \in \mathcal{V}, (i,j) \in \mathcal{E}, \delta_i, \delta_{ij} \in \{0, 1\}.
\end{aligned} \quad (2)$$

This minimization problem can be formulated as a binary integer linear program (BILP), which is NP-complete. To find an exact optimal solution, stock solvers for BILP use branch-and-bound like techniques to intelligently iterate through all possible solutions. However, owing to the NP-complete nature of the problem, it is infeasible to compute the exact optimal solution for many problems of our interest. In fact, in our experiments, the computation time for a branch-and-bound based ILP solver was very high even for toy-sized problems. A standard trick to achieve efficient solutions with some approximation is to use a linear programming (LP) relaxation with rounding to obtain an integer solution. However, in preliminary experiments, the solutions obtained using this approach seemed too inaccurate to be useful. Hence, we approximate this optimization problem as a minimum cost network-flow (MCNF) problem that guarantees us a binary solution in polynomial time. (see supplementary material for more discussion).

## 4. View-graph Selection as MCNF Problem

The network in MCNF problems is a directed graph with at least one source and one sink nodes. We denote the network as $\mathcal{T} = (\mathcal{N}, \mathcal{A})$, where $\mathcal{N}$ represents the set of nodes and $\mathcal{A}$ represents the set of directed edges/arcs. Each edge $(i, j)$ in the network has a cost $c_{ij}$ associated to let across one unit of flow, and the cost incurred by an edge is proportional to the flow ($x_{ij}$) through it. Each edge also has a lower and an upper bound ($l_{ij}, u_{ij}$) on the amount of flow ($x_{ij}$) that can pass through it, known as **capacity constraints**. The source sends a certain units of flow that the sink node must receive. At all other nodes, flow must be conserved, i.e. the total incoming and outgoing flows must be equal. Let us denote the total flow as $F$ and the remainder flow at a node $i$ as $b_i$, then $b_i = F$ when $i$ is source, $b_i = -F$ when $i$ is sink, and $b_i = 0$ otherwise. These constraints are known as **equal flow constraints**. The MCNF problem is about sending the total flow from the source node to the sink node at a minimum cost, without violating the capacity and flow constraints. This minimization with flow and capacity constraints can be described as follows,

$$\begin{aligned}
\text{Minimize} \quad & \text{Subject to,} \\
\sum_{(i,j) \in \mathcal{A}} c_{ij} x_{ij} \quad & l_{ij} \leq x_{ij} \leq u_{ij} \quad \forall (i,j) \in \mathcal{A} \\
& \sum_{\{j:(i,j) \in \mathcal{A}\}} x_{ij} - \sum_{\{j:(j,i) \in \mathcal{A}\}} x_{ji} = b_i \quad \forall i \in \mathcal{N}
\end{aligned} \quad (3)$$

**Network construction** To pose view-graph selection as an MCNF problem, we construct the network as follows. All indicator variables corresponding to selection of views $\{\delta_i\}$, and pairwise geometries $\{\delta_{ij}\}$ are represented using arcs/edges $(i, j) \in \mathcal{A}$ in the network, source and sink nodes are auxiliary. Since view selection variables are represented as arcs, each vertex $i$ in the view-graph corresponds to two nodes, $(2i - 1, 2i)$ in the network. Each odd node $(2i - 1)$ corresponding to the vertex $i$ in the view-graph is connected to the source node and similarly each even node $(2i)$ is connected to the sink node. The arcs corresponding to the pairwise selection variables $\{\delta_{ij}\}$, join the even node of the lower index image with the odd node of the higher index image. This choice prevents cycle formation in the network. Summarizing, the network consists of $|\mathcal{N}| = 2|\mathcal{V}| + 2$ nodes and $|\mathcal{A}| = |3\mathcal{V}| + |\mathcal{E}|$ arcs. These arc connections in the network are summarized in Figure 2 along with a pictorial example.

| Edge Type | Connected Nodes | | Capacity |
|---|---|---|---|
| Source: | $(0, 2i-1)$ | $\forall i \in \mathcal{V}, 0 :$ source | $[0, F]$ |
| Sink: | $(2i, |\mathcal{V}|+1)$ | $\forall i \in \mathcal{V}, |\mathcal{V}|+1 :$ sink | $[0, F]$ |
| Image: | $(2i-1, 2i)$ | $\forall i \in \mathcal{V}$ | $[0, \deg(i)]$ |
| Pairwise: | $(2i, 2j-1),$ | $\forall e(i,j) \in \mathcal{E}, i < j$ | $[0, 1]$ |

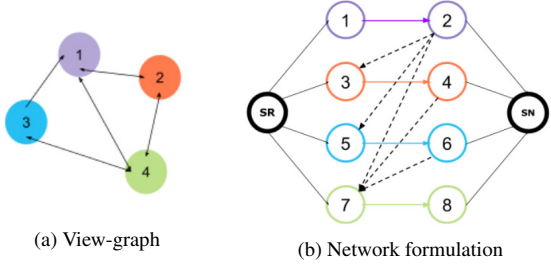

(a) View-graph  (b) Network formulation

Figure 2: The table (top) describes the arc connections in the network based on view-graph vertices and edges. The figure (bottom) shows an example network construction (right), for a sample view-graph (left). Source and sink nodes outlined in black. Image nodes and the corresponding selection arcs are color-coded to match vertices in the view-graph. Pairwise selection arcs are depicted by black dashed lines. (Please see in color.)

**Equal flow and capacity constraints** At source and sink the sent and received flow is equal to the total flow. Equal flow constraints require that in and out flow at every other node remain equal. In our formulation, capacity constraints - lower and upper bounds on flow through an arc are specified based on the edge type as mentioned in Figure 2.

To understand the choice of these capacities, consider the vertex $v_1$ in the depicted view-graph. The arc corresponding to $v_1$'s selection variable $\delta_1$ in the network is $a(1,2)$. Corresponding to $v_1$'s degree in the view-graph, the node $n_2$ in the network has three outgoing arcs $a(2,3)$, $a(2,5)$, and $a(2,7)$ for pairwise selection variables $\delta_{12}$, $\delta_{13}$, and $\delta_{14}$. The flow starting from source node, after passing through $a(1,2)$, should plausibly be able to pass through all three outgoing arcs. Since flow can only be divided in integer units, the minimum capacity of image selection arc $a(1,2)$ has to be at least 3 ($deg(v_i)$). Under minimum cost solution, the flow at any node will continue to take the path of least resistance (cost). Now, suppose that the cost assigned to $a(2,3)$ is the least amongst the three outgoing arcs and its max. capacity is 3 units (or any value $1 < k < deg(v_i)$). In this scenario, $a(2,3)$ being the lowest cost arcs will pull all 3 units of flow from $a(1,2)$, starving the other arcs of any flow and preventing the corresponding view-graph edges from ever getting selected. To avoid this, we restrict the maximum flow through pairwise selection arcs to 1.

**Effect of cost normalization and flow on solution** In minimization sense, negative costs provide encouragement for flow to pass through an arc, whereas positive costs provide discouragement. Suppose, all costs are negative, then total flow $F$ of 1 unit will select the lowest cost chain in the network (often the longest). As we increase the value of total flow, more paths get explored and when $F = |\mathcal{E}|$ all images and EGs get selected. When costs are both encouraging and discouraging, many positive cost arcs will act as barriers for the flow. As a result, at some value of total flow $F$ the selection will (nearly) saturate and may never select the full view-graph. For the proposed applications, we use only encouraging costs ($-1 \leq c_{ij} \leq 0$), while flow remains the only free parameter.

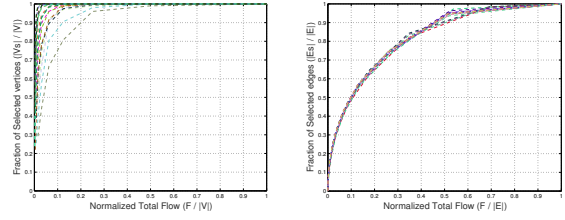

Figure 3: Effect of total flow on vertex and edge selection. It can be seen that the selected vertices and edges increase monotonically (logarithmic) with increase in flow.

**Running time** The main advantage of this formulation is that it can be solved very efficiently. Constructing the network and solving for MCNF takes less than a seconds even for graphs with $\sim$1000 nodes and $\sim$100K edges.

**Tuning flow parameter** To systematically understand the effect of total flow on image and pair selection, we created synthetic view-graphs with varying number of vertices, varying degrees of connectivity simulated using Gaussian assumptions on neighborhood, and randomly assigned (negative) costs (discussed in detail in the suppl. material). We performed sub-graph selection on $\sim$100 such synthetic view-graphs using the proposed MCNF approach for increasing values of total flow and observed a consistent pattern in selection behavior. Figure 3 shows the plots for fraction of total vertices and edges selected vs. normalized total flow. It can be seen that for negative costs, the relation between total flow and selected vertices and edges is logarithmic in nature. This relation is also observed on real-world (Internet) datasets with well-connected view-graphs (see in suppl. material). Depending on the fraction of total vertices and edges we want to be selected, the logarithmic dependence allows us to find the desired sub-graph by a binary search of the flow parameter over the $[0, 1]$ interval with a logarithmic scale. Such an iterative search is particularly effective in view of the extremely low computational time of the MCNF algorithm. In fact, the combined processing time of the the whole search procedure is still an insignificant fraction of the total time required for SfM, and is thus far from being the computational bottleneck.

## 5. Applications and Cost Modeling

To show how the proposed framework can be used to address different concerns using task-specific costs, we tackle two use-cases. We first discuss reconstruction of highly ambiguous datasets with costs designed using specialized priors and later show reconstruction of general SfM datasets using commonly used priors.

### 5.1. Cost modeling for ambiguous datasets

We propose that with conservative selection of an input view-graph such that it consists of a higher fraction of 'true' EGs, correct reconstruction can be recovered without requiring any change in the reconstruction pipeline. To achieve this using the proposed framework, we use three pairwise measures that act as strong priors for disambiguation and express pair selection cost as a linear combination of these three priors (with uniform weights). Note that $c^k(e_{ij})$ denotes the contribution of prior $k$ to the total cost $c$ for the pair $e_{ij}$, and $g$ denotes a normalization function that distributes raw prior value to the desired cost range.

**Context Similarity** Missing correspondences (matches in a pair, not matching the third image in a triplet) are a useful prior for disambiguation. However, by itself, this prior is not very effective for disambiguation and it is commonly used in an inference framework with other priors or with a global objective [32, 16, 12]. Track covisibility statistics [26] are also useful for disambiguation, but it is not straightforward to apply this prior to describe pairwise fitness. In a post-reconstruction disambiguation approach [11], reconstructed 3D points are back-projected into image pairs and conflicting observations in unmatched regions (context) are identified. However, this measure is also not directly useful as our approach is a pre-process to SfM reconstruction. Motivated by these priors, we propose a new, context based pairwise prior that is suitable to our framework.

For image pair $(I_i, I_j)$, the sets of all matched features (matched with any other image) are $S_i$, $S_j$ and the sets of features that match between $(I_i, I_j)$ are $M_i, M_j$. The difference sets $U_i = S_i \setminus M_i$, and $U_j = S_j \setminus M_j$ consist of unique features in $I_i$ and $I_j$. Suppose images $I_i$ and $I_j$ are looking at a scene with duplicate instances of a structure. The features in the match sets $M_i$ and $M_j$ will most likely lie on the duplicate elements. We consider the unique feature sets $U_i$ and $U_j$ to belong to the context regions. If two images are truly looking at the same instance, the context features of both images would have matched similar set of images. On the other hand, if two images are looking at duplicate instances, the context features would be distributed over different sets of images. We find distribution of features in $U_i$ and $U_j$ over all $N$ images in the collection based on their matches and make an $N$ dimensional description of the context space. This concept is illustrated in Figure 4. Context feature of image $I_i$ w.r.t. image $I_j$ can be described as $\mathbf{w}_{ij} = [w_i^1, w_i^2, \ldots, w_i^N] \quad w_i^j = 0$, where $w_i^k = |\{u \in U_i \mid u \leftrightarrow u', u' \in S_k, k \neq j\}|$. This measure is slightly biased against pairs with very low visual overlap, however, combined with the other two measures, it works effectively for selection. We compared the context features of a pair using cosine similarity or hamming distance (after binarization) and found them to be working similarly. The context similarity can be defined as, $c^k(e_{ij}) = g(\mathbf{w}_{ij}^\mathsf{T} \mathbf{w}_{ij})$.

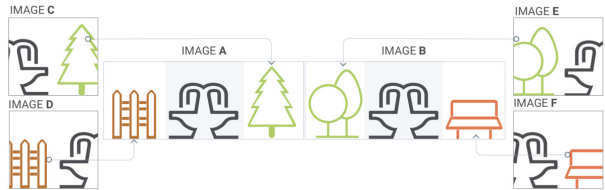

Figure 4: Image pair (A,B) capture a scene with duplicate elements. Though duplicate elements in the images yield many matches, elements in context regions tend to find matches with non-intersecting sets of images.

**Loop consistency** Loop consistency suggests that rotations in an EG triplet when chained should yield identity [7]. We find all triplets in the view-graph and label them as consistent or inconsistent. For each EG, we count the total number of consistent EGs it participates in and use this as a measure of its fitness. Though it is not same as explicitly enforcing loop consistency, it is a convenient way to incorporate a non-local (pairwise) geometric cue as a pairwise prior. $c^k(e_{ij}) = g(\frac{\text{\# consistent triplets on } (i,j)}{median(\text{\# consistent triplets on any}(i,j))})$.

**Multiple motions** We model this prior to disambiguate image pairs capturing repetitive or duplicate instances of some scene elements. We remove the correspondences that satisfy the estimated EG from the initial set of matches and estimate the secondary relative pose using the leftover matches. If sufficient inliers are found, we decompose the relative pose into rotation and translation. We estimate the angular difference between primary and secondary rotations ($\Delta R_\theta$) and also the angle between the both position vectors ($\arccos\left(\Delta C_{ij}^\mathsf{T} \Delta C_{ij}\right)$). If the difference angles are small, the secondary motion is most possibly arising due to threshold sensitivity and measurement drift, otherwise, these indicate presence of correspondences on ambiguous structures. Examples for both these scenarios are shown in Figure 5. $c^k = g(\Delta R_\theta, \arccos\left(\Delta C_{ij}^\mathsf{T} \Delta C_{ij}\right))$. When secondary motion is detected, the value of this prior is very effective in disambiguating (except for the rare instances when camera is purely translating along the direction of the repetition) However, this prior alone is insufficient for disambiguation, as for many incorrect pairs secondary geometry is not detected. We combine this prior with the other two priors only for the pairs where secondary geometry is detected.

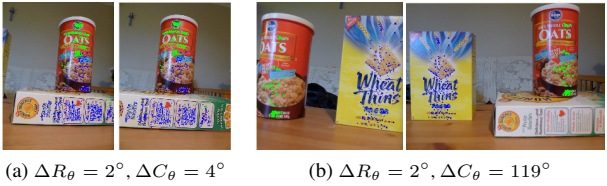

(a) $\Delta R_\theta = 2°, \Delta C_\theta = 4°$     (b) $\Delta R_\theta = 2°, \Delta C_\theta = 119°$

Figure 5: Examples pairs where multiple motions are detected. While for pair (a) the primary and secondary rotations and positions are in agreement, for pair (b) the ambiguity is clearly reflected in position difference.

### 5.2. Cost modeling for general datasets

View-graph selection for general datasets is typically done to achieve complete reconstructions with smaller reprojection errors and shorter run-time. To achieve these goals, we use simple priors based on graph-connectivity and local geometry. These priors are not novel, but instead, they express common-knowledge selection heuristics and criteria of general SfM methods in form of image and pairwise selection costs, suitable for our framework.

**Image selection priors** For the purposes of accurate and complete reconstruction, we consider three image priors, (i) degree of an image in the view-graph, (ii) fraction of an image's features that participates in tracks, (iii) local clustering coefficient of an image. First two measures favor selecting images with many observations and connections to support longer tracks useful for accurate triangulation. Since Internet photo collections often suffer from point of view bias, only connectivity based priors could lead to selections within popular components. To compensate for this effect, we include local clustering coefficient (lcc) of a vertex as an image prior to prefer images that provide connections across components than within components. Image selection cost is a linear combination of these priors with uniform weights.

**Pairwise selection priors** Fitness of a pair in traditional SfM pipelines is measured using two common criteria baseline and number of EG inliers. We use four pairwise priors, (i) number of inliers, (ii) median triangulation angle of pairwise reconstructed features, (iii) overlap (area of a convex hull of the matched features), and (iv) infinite homography, that are reflective of this selection criteria. While the usefulness of first three priors is evident, detecting infinite homography is useful in discouraging pairs with panoramic motion (rotation around a fixed center). Despite a high overlap, such pairs are undesirable as they lead to degenerate or ill-conditioned EG. Simply using homography inliers also rejects valid EGs due to planar regions. We use the fact that calibration normalized infinite homographies are basically rotations [7]. Hence, $H^\mathsf{T}H$ should be close to identity for such pairs and we model this prior as, $||H_{ij}^\mathsf{T} H_{ij} - I||_\mathrm{F}$.

## 6. Results and Discussion

We apply our view-graph (henceforth mentioned as VG for brevity) selection framework for the two use-cases discussed in the previous section and demonstrate results on a variety of datasets. Implementation details of our pipeline are provided in the supplementary material along with additional analysis and results.

### 6.1. Ambiguous datasets reconstruction

With the standard incremental SfM pipeline and VGs selected using our approach, we show successful reconstruction results on a number of highly ambiguous datasets consisting of small-scale laboratory-style scenes [16], and large-scale urban environment scenes [11, 18]. Details of these datasets and selection statistics are given in Table 1.

Figure 6 shows the reconstruction results for small-scale ambiguous scenes [16]. It can be seen that with selected VGs, we are able to recover true structures for all datasets. To highlight the effectiveness of disambiguation specific priors, we also reconstruct these datasets for VGs selected using random costs (fails on all datasets), and baseline priors based costs (fails on all but the 'Books' dataset). Figure 7 shows the reconstruction results for large-scale urban ambiguous scenes [11, 18]. Method proposed in [11] operates post-reconstruction to split incorrectly merged model parts and takes ∼16 to ∼85 minutes to process for these datasets. Our framework pre-selects the view-graph and reconstruction is performed without any additional processing. Recently proposed method of [30] also tackles the disambiguation problem as a pre-process to SfM but their method takes 2–11 minutes on these datasets. Our subgraph selection framework is extremely efficient taking 1–2 seconds for constructing and solving the MCNF problem for these and even larger datasets. Moreover, our framework is intended to be general purpose with disambiguation as one of the specific objectives. Result for 'ANC' dataset is not shown as it is incomplete as compared to [11, 30].

**Criteria for flow parameter search** As can be seen in Table 1, for the lab-style datasets, the full view-graph is generally quite dense, with an average vertex degree (edges per vertex) ∼32% of total vertices. However, due to the very high ambiguity in the scenes, large number of pairs are expected to be outliers. Therefore, it is makes intuitive sense to expect that a pruned view-graph comprising a reasonably small fraction (∼25%) of the total number edges would lead to a good reconstruction. On the other hand, the VGs for urban ambiguous scenes [11] are not well-connected, as these datasets are already sampled subgraphs (iconics) of the original image set. This is also evident from the fact that average number of edges per vertex is only ∼5% of total vertices. In view of this sparse connectivity, it makes sense to keep as many vertices as possible and a bigger frac-

| Sr. | Dataset | $|\mathcal{V}|$ | $|\mathcal{E}|$ | F | $|\mathcal{V}_s|$ | $|\mathcal{E}_s|$ | $t_{sel}$ |
|---|---|---|---|---|---|---|---|
| 1 | Cereal | 25 | 228 | 4 | 25 | 49 | 0.0137 |
| 2 | Cup | 64 | 1217 | 16 | 64 | 265 | 0.0268 |
| 3 | Oats | 24 | 220 | 4 | 24 | 49 | 0.0137 |
| 4 | Street | 19 | 95 | 2 | 19 | 26 | 0.0089 |
| 5 | Books | 21 | 161 | 4 | 21 | 41 | 0.0135 |
| 6 | Desk | 31 | 261 | 4 | 31 | 64 | 0.0138 |
| 7 | ANC | 448 | 5037 | 512 | 416 | 2497 | 0.1288 |
| 8 | ADT | 381 | 3627 | 512 | 340 | 2148 | 0.0940 |
| 9 | RDC | 271 | 3378 | 128 | 258 | 1498 | 0.0717 |
| 10 | BG | 161 | 2003 | 128 | 129 | 848 | 0.0522 |
| 11 | CSB | 277 | 5191 | 128 | 233 | 1380 | 0.0743 |
| 12 | ToH | 341 | 50332 | 32 | 341 | 1990 | 0.2372 |

Table 1: Details of ambiguous datasets and selection statistics. $t_{sel}$ shows the combined running time for all search iterations of MCNF solver. Abbreviated labels correspond to these datasets : ANC – Alexander Nevsky Cathedral, ADT – Arc de Triomphe, BG – Brandenburg Gate, CSB – Church on Splilled Blood, RDC – Radcliff Camera, ToH – Temple of Heaven.

| Dataset | VG | $|\mathcal{V}|$ | $|\mathcal{E}|$ | Incremental SfM $R_{err}$ | $T_{err}$ | $r_{err}$ | Global SfM $R_{err}$ | $T_{err}$ | $r_{err}$ |
|---|---|---|---|---|---|---|---|---|---|
| Castle | S | 30 | 49 | 2.44 | 0.15 | 0.38 | 2.21 | 1.29 | 1.01 |
|  | F |  | 118 | 2.22 | 0.22 | 0.34 | 2.17 | 7.49 | 1.14 |
| Fountain | S | 11 | 21 | 2.90 | 0.01 | 0.29 | 2.82 | 0.29 | 0.35 |
|  | F |  | 25 | 2.90 | 0.01 | 0.72 | 2.82 | 0.27 | 0.59 |
| Herzjesu | S | 25 | 55 | 2.36 | 0.03 | 0.50 | 2.38 | 0.75 | 1.18 |
|  | F |  | 128 | 2.38 | 0.02 | 0.43 | 2.39 | 0.56 | 1.71 |

(a) MVS dataset statistics

| Dataset | VG | $|\mathcal{V}|$ | $|\mathcal{E}|$ | $t_{sel}$ | $N_c$ | $r_{err}$ | $R_{err}$ | $T_{err}$ | $t_{sfm}$ |
|---|---|---|---|---|---|---|---|---|---|
| Notre Dame | S | 659 | 16970 | 1.744 | 628 | 1.41 | 0.072 | 0.195 | 1151 |
|  | F | 714 | 46746 | – | 682 | 1.53 | 0.089 | 0.217 | 1760 |
| Pantheon | S | 761 | 15975 | 3.721 | 754 | 1.06 | 0.098 | 0.310 | 1785 |
|  | F | 781 | 139630 | – | 775 | 1.31 | 0.125 | 0.309 | 3601 |
| St. Peters | S | 1132 | 39640 | 2.864 | 1095 | 1.341 | 0.037 | 0.517 | 1147 |
|  | F | 1155 | 119977 | – | 1111 | 1.458 | 0.028 | 0.496 | 1367 |

(b) Internet landmarks datasets statistics

Table 2: VG selection and reconstruction statistics for MVS benchmark and Internet landmarks datasets. Labels 'S' and 'F' indicate selected and full VGs. $|\mathcal{V}|$, $|\mathcal{E}|$ indicate the vertices and edges in associated (selected or full) VGs. $N_c$ is the number of reconstructed cameras and $R_{err}$, $T_{err}$, and $r_{err}$ indicate median rotation, translation, and reprojection errors. $t_{sel}$ and $t_{sfm}$ indicate the time taken for MCNF based VG selection (cummulative) and SfM reconstruction respectively.

tion of edges. Practically, we chose these threshold to be, $|\mathcal{V}_s| >= 80\%$ of $|\mathcal{V}|$, and $|\mathcal{E}_s| >= 5|\mathcal{V}|$ and use the method described in section 4 to get the desired sub-graph by efficiently searching for the appropriate flow value.

### 6.2. General datasets reconstruction

We show that the proposed framework is versatile and can also be used for the general goals of accuracy and efficiency for reconstruction of standard SfM datasets, by modeling image and pair selection costs with well-known and commonly used SfM priors. For this task, we show the results on two datasets, *(i)* MVS benchmark [23], and *(ii)* Internet landmarks. These datasets are reconstructed with incremental [29] and global SfM [24] pipelines, using both full VGs and VGs selected by our method.

MVS benchmark consists of three toy-sized datasets with ground-truth (GT) camera positions. Table 2a shows that, for both SfM methods, the selected VGs based reconstructions are comparable to the full VGs based reconstructions. Flow parameter for these selections was chosen such that all vertices are selected (to enable GT comparisons).

For large-scale Internet landmarks datasets, we obtain reconstructions for selected VG and full VG with global SfM method, and in absence of ground-truth, compare these with the full VG based reconstructions obtained with incremental SfM. Table 2b shows the selection and reconstruction statistics for these datasets. For completeness of the recovered structure, it is desirable to have as many vertices as possible in the subgraph. For efficiency, it is desirable to select fewer edges, however too few edges (low vertex degree) can lead to many short feature tracks and high reprojection errors. With these considerations, we keep $|\mathcal{V}_s| = 90\%$ of $|\mathcal{V}|$ and $|\mathcal{E}_s| = 20|\mathcal{V}|$ as flow search criteria. Table 2b shows that the reconstructions for selected VGs are comparable or more accurate as compared to the full VG reconstructions and SfM with selected VGs take notably shorter time. Qualitative results are provided in the supplementary material.

### 7. Conclusions and Future Work

We presented a novel and efficient, unified framework for selecting subgraphs from initial view-graphs that can achieve different selection objectives with appropriately modeled image and pairwise selection costs. This mechanism provides an interesting way to separate dataset and task specific challenges from the standard SfM pipeline, thereby improving its generality. We demonstrated utility and potential of this framework by achieving satisfactory results for two different objectives with task-specific cost modeling. One interesting way to achieve even further abstraction to this problem would be to replace hand-designed costs by a weighted combination ($c_i = \sum_k \alpha_k f_k(i)$) of a number of known and designed priors. Cost formulation of this form would be expressive enough to cater to a wide variety of selection objectives. The problem of modeling costs to meet the desired objective then translates to that of devising new priors to add to the combination and finding the right weights for prior combination. In future, we wish to explore this direction for extending our framework. While this is non-trivial, it can lead to interesting directions of research for searching/learning new priors and the combination weights.

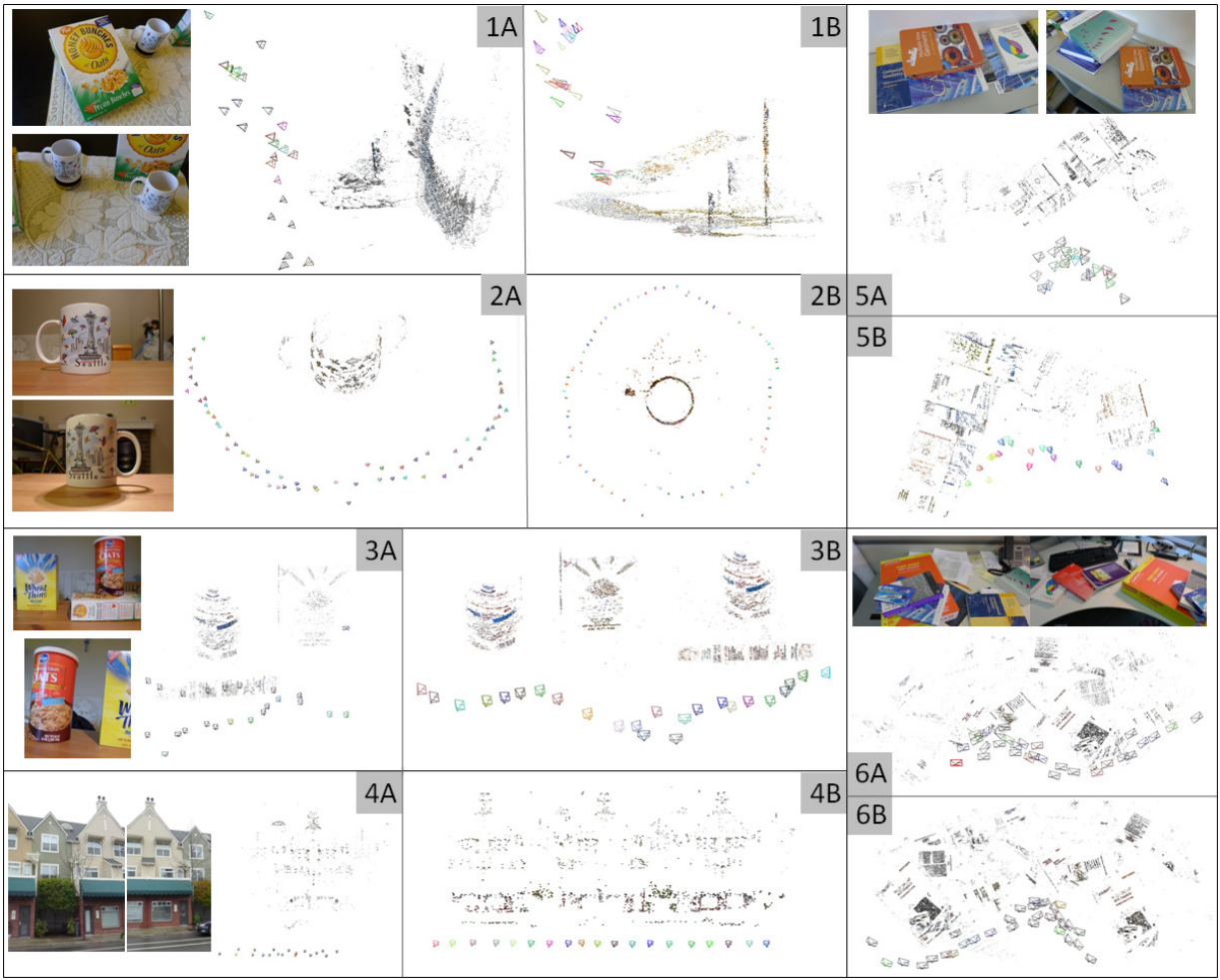

Figure 6: Reconstructions for small ambiguous datasets (numbered as per Table 1): (A) full VG, (B) selected VG.

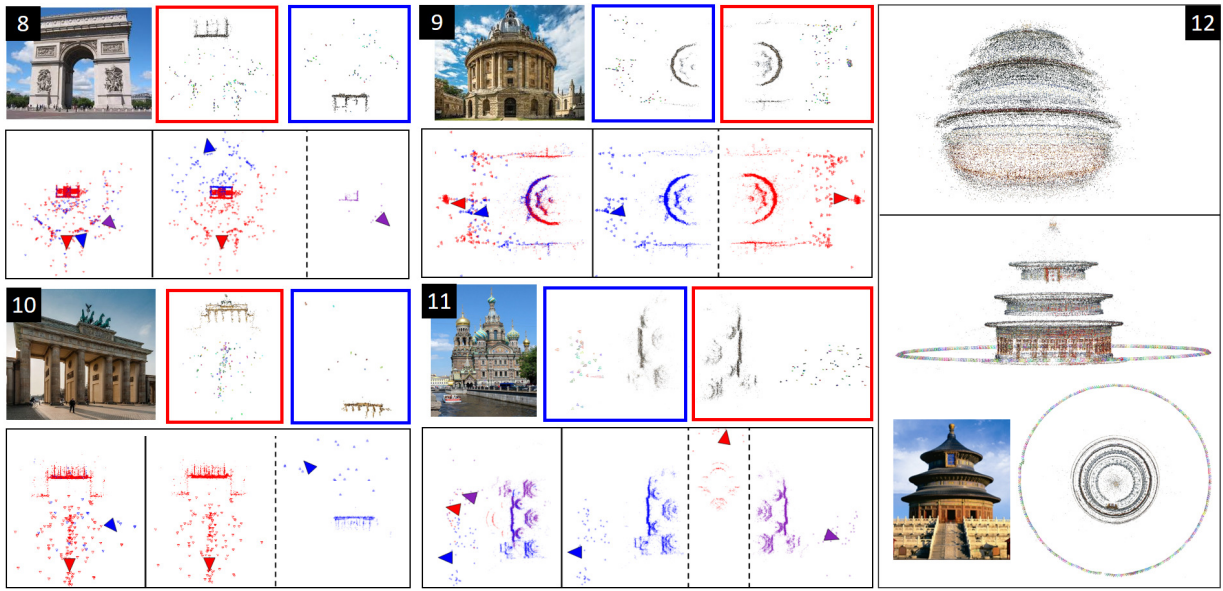

Figure 7: Reconstructions for large ambiguous scenes (numbered as per Table 1). For 8 to 11, bottom left – incorrect model with full VG, bottom right – result of [11], and top row – our result (color-coded to match splits of [11]). For 12, top – full VG, bottom – ours.

# A Unified View-graph Selection Framework for Structure from Motion

Supplementary Material

## Summary

This supplementary document provides information about the following. (i) Implementation details. (ii) Explanation of the relationship between the selection problem (eqn. 2) and the MCNF problem (eqn. 3) discussed in the paper. (iii) Explanation of synthetic view-graph based experiments conducted to study the effect of flow parameter on selection. (iv) Qualitative results (point clouds) and analysis of selection and reconstruction statistics for Internet Landmarks dataset [6, 4]. (v) Ablation study to evaluate effectiveness of disambiguation priors.

## I. Implementation details

Our pipeline is implemented in C++ with necessary bindings. We use David Lowe's implementation[1] for SIFT feature detection and descriptor computation, and approximate near neighbour (Kd-tree based) approach[2] with ratio-test for feature matching. These components are standard in SfM community. We use `Theia` SfM library[3] for relative pose estimation, geometric verification of matches, and for computing geometry based (pairwise and tripletwise) priors. For MCNF based optimization, we use `mosek` optimization library[4] with C++ interface. The optimization is done efficiently using the Network Simplex method.

All disambiguation results use incremental SfM (VisualSfM) based reconstructions. Results on Landmarks datasets use global SfM based reconstructions as we want to compare accuracy of these reconstructions with respect to incremental SfM based baseline reconstructions. We use `Theia`'s implementations of robust rotation averaging [3] and 1D translation estimation [7] as global SfM pipeline.

All CPU based experiments (selection and global SFM reconstructions) are run single-threaded on a machine with Intel Xeon 2.50GHz processor and 12GB RAM. For VisualSfM based reconstructions, we use a similar machine with NVidia 970 GPU.

We will make our codes for prior computation, cost modeling, and network based optimization, publicly available along with the paper.

## II. Reducing selection problem to MCNF

Network flow problems have been routinely used to solve selection problems in the past. The project selection problem is a typical example.[5] In the case of project selection, there are a set of $P = \{v_i\}$ possible projects, and unary costs associated with selecting each project $p_{v_i}$. There are also a set of prerequisistes, where a project $v_j$ cannot be executed unless a project $v_i$ is also executed. The project selection problem is the task of chosing a *feasbile set* of projects $A \subseteq P$ such that the costs (unary and pairwise) are maximized. It can be solved as a maximum flow problem. A directed graph is constructed by assigning a node to each project. A project with positive revenue is connected to the source $s$ of the graph, and a project with negative revenue is added to the sink $t$ of the graph. Projects that are dependent are connected by a directed edge $e(i,j)$ if executing project $v_j$ needs project $v_i$ to be executed first. Edges $(s, v_i)$ have flow capacity $p_{v_i}$ if $p_{v_i} > 0$ and edges $(e, v_i)$ have flow capacity $p_{v_i}$ if $p_{v_i} < 0$. Edges $(v_i, v_j)$ have infinite capacity if $v_i$ is a prerequisite for $v_j$. Proof of their equivalence can be found in literature.[6]

Note that in our case, we are interested in a slightly different goal, and hence we use a minimum cost network flow problem. Firstly, in our case, there are no prerequisites. Thus our flow capacities corresponding to pairwise terms would be set to non-prohibitive values. However, this can result in a solution where all capacities of selected edges are not maxed out, which is undesirable in our case.

**Minimum Cost Network Flow (MCNF)** This is where minimum cost network flow comes into the picture. We know that the *total unimodularity* [1] property of MCNF problems allows us to get integer solutions, when all the capacities are integers. Thus, we convert the above graph by constraining the pairwise selection edge capacities to be 1 (indicating whether the current edge is selected or not). The cost of each selection is recast as the cost of passing flow through each edge. Finally, the unary terms in equation 2 can be cast as edges by splitting a node in two and adding an edge between them. In such a case, only one of the two nodes would be connected to the source (or sink) and the other node will be used to connect to all the pairwise edges. This ensures that if one node is connected to all edges that *receive* a flow, and only the other node is connected to all edges that *transfer* flow to other nodes. The constraints added in equation 2 to ensure that pairwise selection is not independent of image selection are implicitly modeled by the network construction design.

---

[1] http://www.cs.ubc.ca/~lowe/keypoints/
[2] https://www.cs.umd.edu/~mount/ANN/
[3] http://www.theia-sfm.org/
[4] https://www.mosek.com/
[5] https://en.wikipedia.org/wiki/Max-flow_min-cut_theorem#Project_selection_problem
[6] https://courses.engr.illinois.edu/cs473/sp2013/w/lec/19_add_notes.pdf

Thus the entire problem is now cast as a problem of passing flow $F$ through an MCNF network while minimizing cost. The fact that the total flow $F$ is a free variable is discussed in Section 5 of the paper. Further examples of recasting selection as an MCNF problem can be found in multi-target tracking in videos [2] and matching data association in SLAM [5].

## III. Experiments with synthetic view-graphs

To systematically study the effect of flow parameter on sub-graph selection, we solve the proposed MCNF problem for a large number of synthetic graphs. The results of this experiment are shown and discussed in the main paper (Section 4, Figure 3) . We construct the synthetic graphs to simulate connectivity of real-world unorganized image collections using the below process,

1. Begin with the number of vertices ($N$) in the graph and average degree ($d_{avg}$) as a fraction of $N$.

2. The degree of each vertex is sampled from a Gaussian distribution with $\mu = d_{avg}$ and varying $\sigma$.

3. The adjacency list of each vertex is also created by sampling a Gaussian distribution. For a vertex $i$, the $d_i$ connected vertices are sampled from a Gaussian distribution with $\mu = i$ and $\sigma$ as a function of average degree ($d_{avg}$). When vertex $i$ is sampled, we randomly choose between vertices $i-1$ and $i+1$ for ensuring validity of edges.

4. The vertices are assumed to be arranged, in a linear fashion (e.g, capturing a facade or a street), and in a loop (e.g, capturing around a building or a block). For loopy configuration, we use modulo for defining vertex adjacency.

5. After defining the connections, each edge is randomly assigned a cost between $-1$ and $0$.

6. Finally, the vertices are shuffled randomly to simulate the unorganized nature of SfM view-graphs.

We vary the number of vertices from 100 to 5000 in multiples of 2, the average degree from 5% to 40% in multiples of 5, and loopy/linear arrangement to obtain synthetic graphs with a wide range of scale and connectivity. In most Internet collection view-graphs, many images come from popular viewpoints; these images have higher degree in the graph and are more likely connected to other adjacent images. While images corresponding to less popular viewpoints also exist in the datasets with fewer connections to popular viewpoint images. This scenario is well captured by our Gaussian sampling and neighbourhood assumptions in our simulation. Hence, the synthetic view-graphs are good approximations of connectivity in real-world unorganized community photo collections. In Figure 2, we show the effect of varying total flow on selection statistics. It can be seen that observations on Internet landmark datasets are similar to that of the synthetic graph experiments (Figure 3 in the main paper).

## IV. Internet landmarks reconstructions

We used our VG selection framework on three large unorganized Internet datasets of popular landmark sites (statistics given in the main paper, Section 6.2, Table 2). We reconstruct the scenes using selected view-graphs and full view-graphs with global SfM pipeline (typically slightly less robust than incremental SfM methods) in order to compare the reconstruction accuracy w.r.t. incremental SfM based baseline reconstructions. To understand the effect of total flow parameter on selection and reconstruction, we plot the selection and reconstruction statistics for increasing value of total flow parameter in Figure 2.

The quantities $|\mathcal{V}_s|/|\mathcal{V}|$, $|N_c|/|\mathcal{V}|$, and $|\mathcal{E}_s|/|\mathcal{E}|$ indicate the fraction of total vertices selected, fraction of total vertices reconstructed, and fraction of total edges (EGs) selected. The rotation and translation errors ($R_{err}$ and $T_{err}$) are divided by max. values to bring these in the same range. It should be observed that fraction of reconstructed vertices remain close to the fraction of selected vertices at all levels of selections. This result indicates that the selection scheme naturally endorses connectivity. This is a result of using only encouraging costs for selection that prefers to select longer chains over disjoint segments. Also most vertices are reconstructed and the reconstruction errors are stable at much lower value of flow (and selected pairs) than full VG selection. The flow value is on a logarithmic scale for a better depiction. It can be also seen that for all three datasets the pair selection vs. flow curve behaves very similarly to the synthetic view-graphs. This result validates our choice of using synthetic view-graph based selection statistics to determine the flow parameter search strategy in practice. The selected VG reconstructions are qualitatively similar or better than full VG reconstructions (shown in Figure 1).

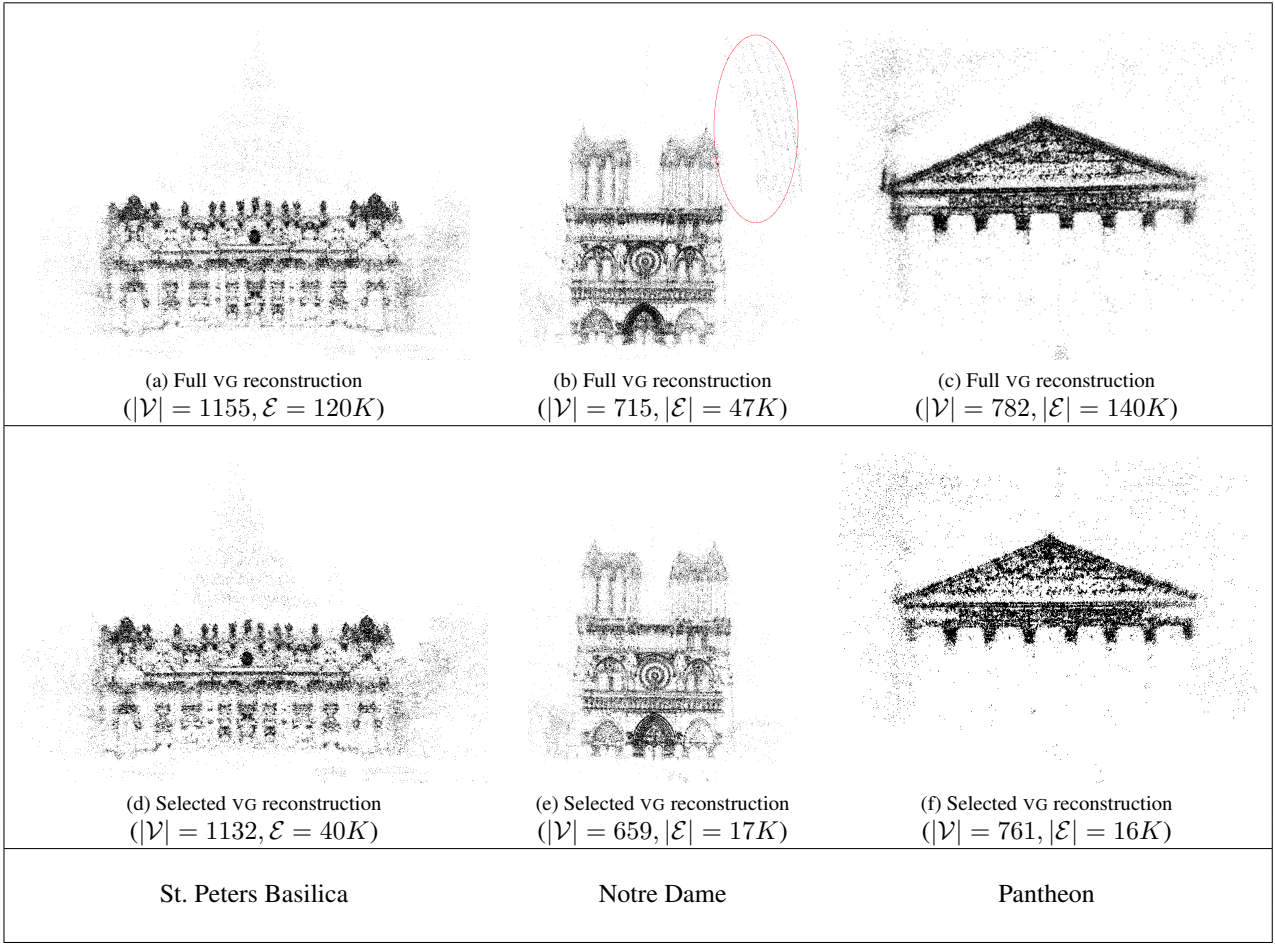

Figure 1: Renders of point clouds for full VG and selected VG reconstructions of Internet landmarks datasets. Notice the ghosting in Notre Dame reconstruction with full view-graph (highlighted by a red oval).

Subfigure labels:
(a) Full VG reconstruction ($|\mathcal{V}| = 1155, \mathcal{E} = 120K$)
(b) Full VG reconstruction ($|\mathcal{V}| = 715, |\mathcal{E}| = 47K$)
(c) Full VG reconstruction ($|\mathcal{V}| = 782, |\mathcal{E}| = 140K$)
(d) Selected VG reconstruction ($|\mathcal{V}| = 1132, \mathcal{E} = 40K$)
(e) Selected VG reconstruction ($|\mathcal{V}| = 659, |\mathcal{E}| = 17K$)
(f) Selected VG reconstruction ($|\mathcal{V}| = 761, |\mathcal{E}| = 16K$)

St. Peters Basilica | Notre Dame | Pantheon

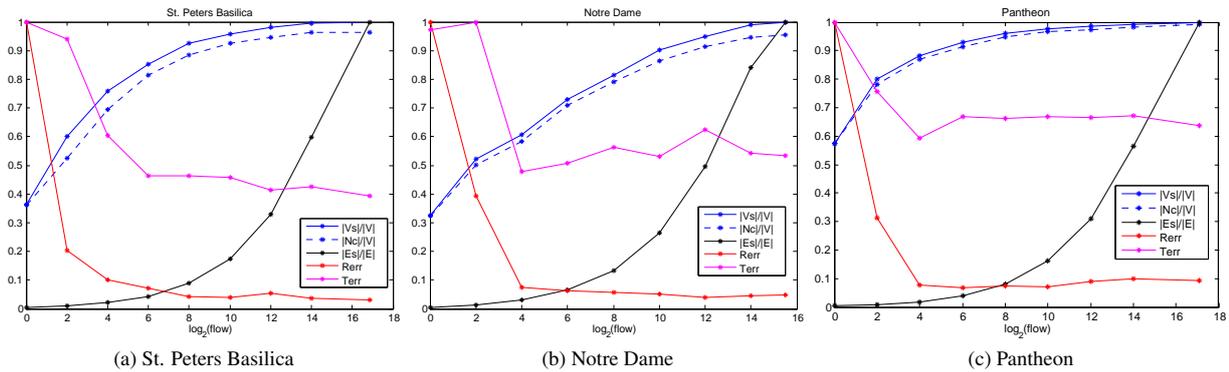

(a) St. Peters Basilica  (b) Notre Dame  (c) Pantheon

Figure 2: View-graph selection and reconstruction statistics with increasing values of flow for three datasets. The quantities $|V_s|/|V|$, $|N_c|/|V|$, and $|E_s|/|E|$ indicate the fraction of total vertices selected, fraction of total vertices reconstructed, and fraction of total edges (EGs) selected. The rotation and translation errors ($R_{err}$ and $T_{err}$) are divided by max. values to bring these in the same range. These plots are explained in section IV.

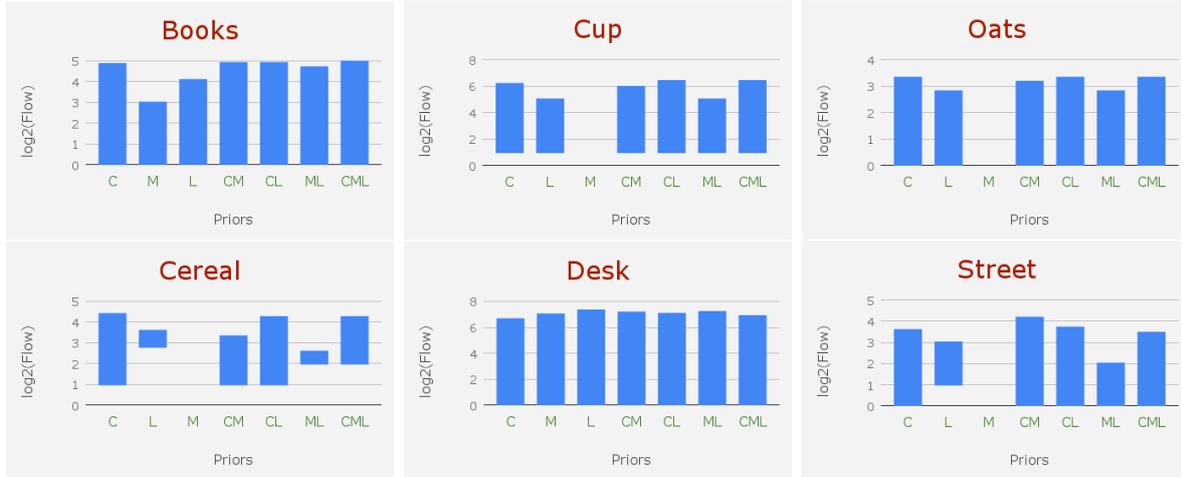

Figure 3: Range of flow values (shown on $y-$axis) for which given prior combination (shown on $x-$axis) leads to correct reconstructions. Priors that lead to correct reconstruction for higher values of flow are better at disambiguation and more robust, since this behavior indirectly indicates selection of fewer outlier EGs. For most datasets, context prior performs better than loop prior. Combination of all priors performs equivalently for practical purposes.

## V. Ablation study of disambiguation priors

In the paper, we proposed three pairwise priors for disambiguation: (i) context similarity (C), (ii) loop consistency (L), and (iii) multiple motions (M). The results shown in the paper are obtained by using a (uniformly) weighted combination of all three priors to model pairwise selection costs. Here, we show the effectiveness of each prior and pairwise combinations on selection and in turn the reconstruction.

Table 1 shows correctness of reconstructions for VGs selected with costs modeled using individual disambiguation priors and their possible combinations (for a fixed value of flow). Figure 4 and Figure 5 shows the qualitative reconstruction results for these selections. It can be seen that selection with context similarity prior is effective to recover all structures correctly. Loop prior is effective for all but 'Cereal' dataset. For this set, with only loop prior, the second instance of the cup (in the center) is not reconstructed (highlighted by red in qualitative result). As mentioned in the main paper, multiple motions are not always detected for all pairs. Hence, standalone this prior is very weak and fails to recover four out of six datasets. As shown in the qualitative results, using only multiple motions based prior leads to severe ghosting errors for 'Oats', 'Street', 'Cereal' reconstructions and fails to recover any model for 'Cup' dataset.

For the chosen values of flow (indicated by $\mathcal{F}$ in Table 1), all combinations (CM, CL, LM, CLM) lead to correct reconstructions of all datasets. In order to further evaluate the effectiveness of the priors and the prior combinations, we reconstruct these datasets using the VGs selected with each prior (and combination) for increasing value of flow. As explained in the main paper, increase in flow value leads to increase in total number of selected pairs (and in turn, selected number of outlier pairs). A good disambiguation prior (combination) is the one that maintains a small ratio of selected outlier pairs to total selected pairs. In other words, the prior combination that works for a higher range of flow (and hence a higher number of selected EGs) is more effective at disambiguation. This analysis is shown in Figure 3. It can be seen that among individual priors (C, L, M), context prior is clearly the most effective at disambiguation for all datasets. Among combinations, CL, CM and CLM perform equivalently, and are marginally better than context prior alone for some datasets.

|        |    | \multicolumn{7}{c}{Priors and Combinations} |
|--------|----|----|----|----|----|----|----|----|
|        | $\mathcal{F}$ | C | L | M | CL | CM | LM | CLM |
| Books  | 4  | ✓ | ✓ | ✓ | ✓ | ✓ | ✓ | ✓ |
| Cereal | 4  | ✓ | ✗ | ✗ | ✓ | ✓ | ✓ | ✓ |
| Cup    | 16 | ✓ | ✓ | ✗ | ✓ | ✓ | ✓ | ✓ |
| Desk   | 4  | ✓ | ✓ | ✓ | ✓ | ✓ | ✓ | ✓ |
| Oats   | 4  | ✓ | ✓ | ✗ | ✓ | ✓ | ✓ | ✓ |
| Street | 2  | ✓ | ✓ | ✗ | ✓ | ✓ | ✓ | ✓ |

Table 1: Correctness of reconstruction for VG selected with pairwise costs based on various combinations of disambiguation priors (for flow value indicated by $\mathcal{F}$). It can be seen that individually context prior is the most effective. Qualitative results for these reconstructions are shown in Figure 4 and Figure 5.

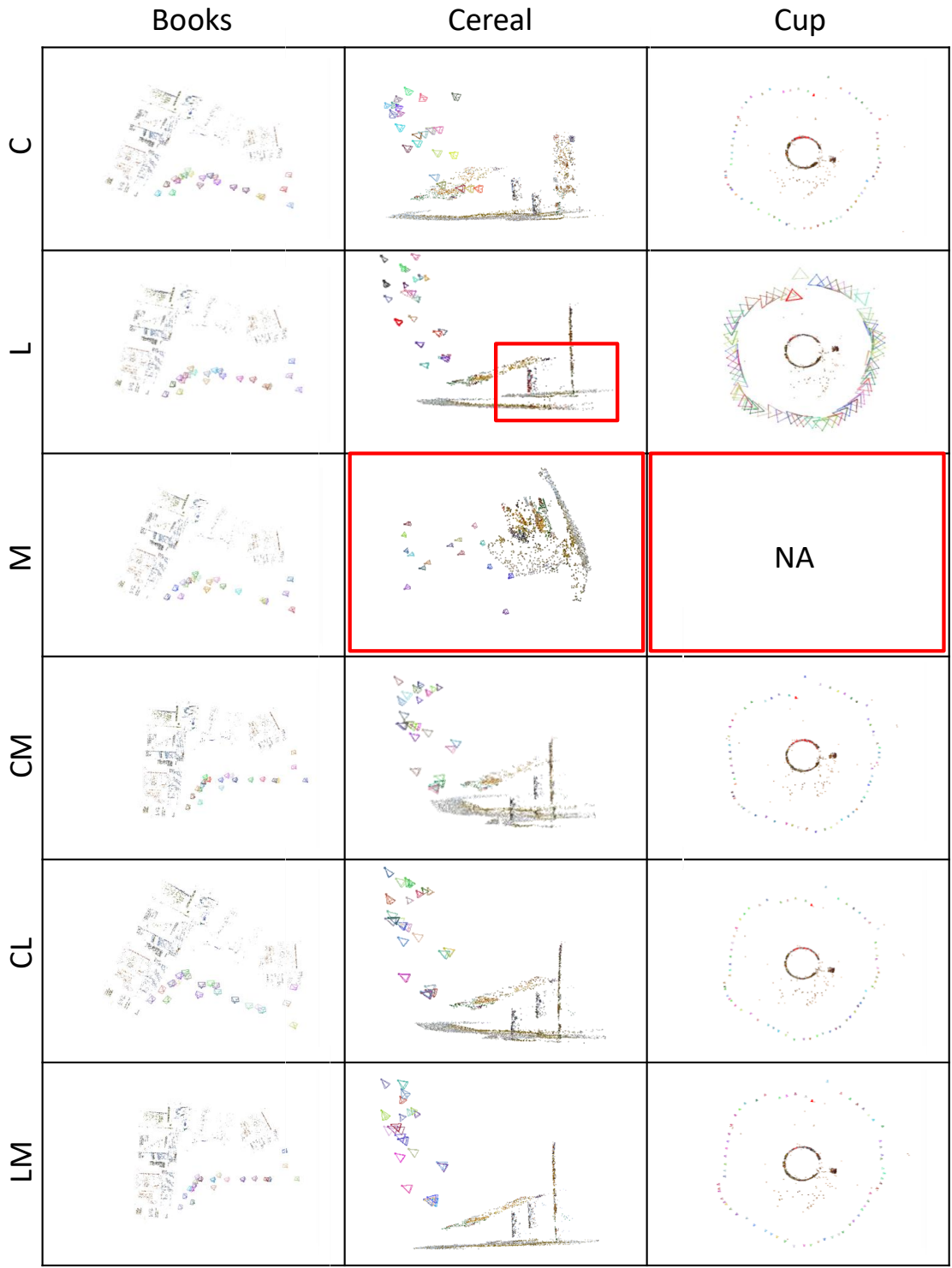

Figure 4: Priorwise qualitative results for three datasets. Wrong reconstructions are highlighted by red boxes.

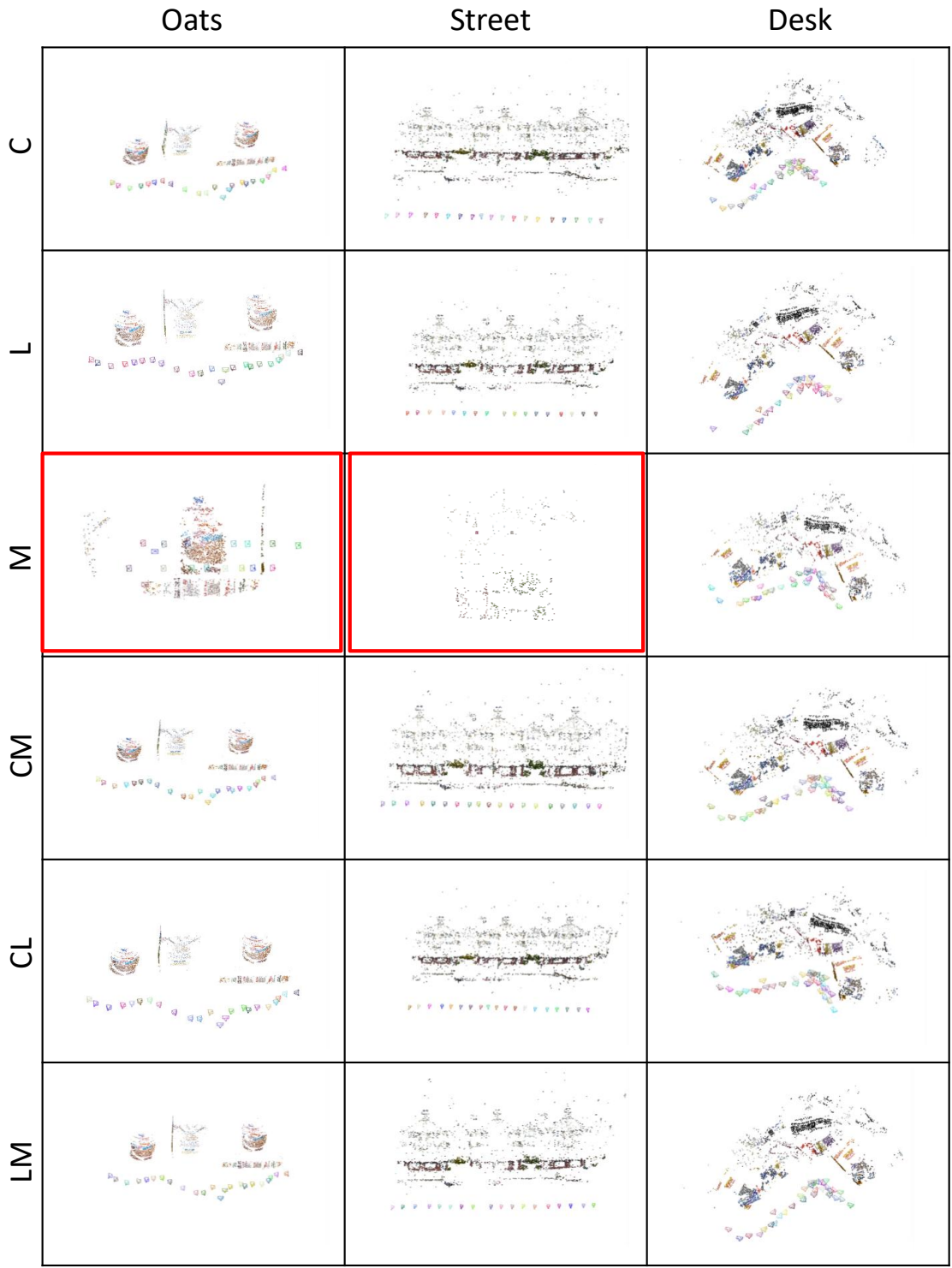

Figure 5: Priorwise qualitative results for three datasets. Wrong reconstructions are highlighted by red boxes.